\documentclass{article}

\usepackage[square,sort,comma,numbers]{natbib}

\usepackage[final]{nips_2018}

\usepackage[utf8]{inputenc} 
\usepackage[T1]{fontenc}    
\usepackage{hyperref}       
\usepackage{url}            
\usepackage{booktabs}       
\usepackage{amsfonts}       
\usepackage{nicefrac}       
\usepackage{microtype}      
\usepackage{enumitem}
\setlist{nolistsep}

\title{Explaining Explanations to Society}

\author{
  Leilani H. Gilpin\\
  MIT CSAIL \\
  \texttt{lgilpin@mit.edu}
  \And 
  Cecilia Testart\\
  MIT CSAIL \\
  \texttt{ctestart@mit.edu}
  \And 
  Nathaniel Fruchter\\
  MIT CSAIL \\
  \texttt{fruchter@mit.edu}
  \And 
  Julius Adebayo \\
  MIT CSAIL \\
  \texttt{juliusad@mit.edu}}

\begin{document}

\maketitle

\begin{abstract}
There is a disconnect between explanatory artificial intelligence (XAI) methods and the types of explanations that are useful for and demanded by society (policy makers, government officials, etc.) Questions that experts in artificial intelligence (AI) ask opaque systems provide \emph{inside} explanations, focused on debugging, reliability, and validation. These are different from those that society will ask of these systems to build trust and confidence in their decisions. Although explanatory AI systems can answer many questions that experts desire, they often don't explain why they made decisions in a way that is precise (true to the model) and understandable to humans.  These \emph{outside} explanations can be used to build trust, comply with regulatory and policy changes, and act as external validation. In this paper, we focus on XAI methods for deep neural networks (DNNs) because of DNNs' use in decision-making and inherent opacity. We explore the types of questions that explanatory DNN systems can answer and discuss challenges in building explanatory systems that provide outside explanations for societal requirements and benefit.
\end{abstract}

\section{Introduction}
There has been a recent surge of work in explanatory artificially intelligent (XAI) systems. One reason these systems are gaining traction is due to changes in policy, law, and regulation. Indeed, with the rise of AI-based decision making in areas of societal interest--from finance and employment to driving and journalism--policymakers see the need to discuss certain standards around XAI.

For example, the European Union’s General Data Protection Regulation (GDPR) creates obligations for automatic decision making processes \cite{gdpr}, with a provision including right to explanation. This broad obligation puts the burden on those who process data with (and develop for) AI systems to generate reasonable explanations for their systems’ decision making processes. We can also observe broad societal concerns with issues of AI liability \cite{liability}. For example, as autonomous vehicles are being introduced, we need a better understanding of what happened in the case of an accident (as it has already happened \cite{uber}). How can there be an appropriate investigation process if opaque decision-making algorithms are involved?  How can we ensure that these machines are acting in our best interest?

AI algorithms and more general, complex AI systems cannot currently provide answers to these prior questions.  These algorithms and systems are not built to explain to the general public nor policy-makers. Although there have been calls for work on creating systems and algorithms that can interpret \cite{finale, rudin-interpret} and explain \cite{gunning-xai} some parts of their decisions, the current state-of-the-art explanatory systems are made for the \emph{programmer} or expert, not an end user or policy-maker.  The key difference here is that the current systems produce what we refer to as \emph{inside} explanations.  They point to a plausible technical explanation, either by looking at the relationships between the inputs and outputs of a model, or examining the role of individual parts, or by producing a surface-level explanation itself. Crucially, though, these explanations do not answer  \emph{why} questions. Continuing with the autonomous vehicle example, when an accident happens involving this autonomous machine, police officials, insurance companies, and the people who are harmed will want to know who or what is accountable for the accident and why it happened.

In this paper, we examine the types of questions that explanatory DNN algorithms can and cannot answer.  We focus on DNNs specifically because of the recent shift in AI  research from symbolic approaches to machine learning and deep learning \footnote{http://www.aiindex.org/2017-report.pdf}, and because these are the systems are making safety-critical decisions in applications like autonomous driving\cite{bojarski2016end} and malware detection\cite{malware}.  In order to bridge the gap between the current, technical deep neural network explanations (which we refer to as \emph{inside} explanations) and the explanations that answer \emph{why} questions that would benefit society (which we refer to as \emph{outside} explanations) we must develop explanations for DNNs that can answer these questions and be probed.  We extend the work of a previously defined \cite{review} taxonomy of explanations by looking at the specific \emph{questions} each class can and cannot answer, and stress the necessity and technical challenges for these systems to be probed and answer why questions.  We motivate future work on bridging the gap between current explainable methods by incorporating the types of questions and explanations society would like to know.

By doing this, we attempt to  ``bridge the gap'' between current, technical deep neural network explanations and explanations that answer \emph{why} questions beneficial to society. We approach this task in four main ways: 

\begin{itemize}
    \item We differentiate between \emph{inside} (technical) and \emph{outside} (\emph{why}) explanations.
    \item We extend the work of a previously defined \cite{review} taxonomy of explanations by looking at the specific \emph{questions} each class can and cannot answer.
    \item We discuss the necessity of probing of AI systems and the technical challenges inherent in creating \emph{outside} explanations.
    \item We motivate future work on bridging the gap between current explainable methods by incorporating the types of questions and explanations society would like to know.
\end{itemize}

\section{Related Work}
In this work, we focus on the types of questions and explanations that explanatory DNN methods can answer.  Recent work has looked at ways to correct neural network judgments \cite{judements} and different ways to audit such networks by detecting biases \cite{audit}. But these judgments are not enough to completely understand the model's decisions-making. Other work answers why questions by finding similar data points \cite{lookslike}. Although these methods are clearly interpretable, they do not provide any unique insights into why the model made those decisions.  Other work examining best practices for explanation \cite{sheh2018defining} provides a set of categories, but does not evaluate the questions that explanatory systems should be able to answer; which is necessary for policy makers and societal trust in DNN decision processes.  

Since we are interested in the societal expectation of explanations, it is important to examine prior initiatives on the legal side. The desire for explanations in certain sectors is not new. For example, the U.S. Fair Credit Reporting Act creates obligations for transparency in certain financial decision-making processes, even if they are automated \cite{landau}. The role of explanation has been examined to enforce accountability under the law \cite{ai-ethics}.  Similar recommendations in using explanations in law have been examined in promoting ethics for design \cite{saving}, for privacy \cite{mulligan}, and liability for machines \cite{liability}.

In this paper, we explicitly examine DNNs, even though there have been important developments in explanatory systems not tailored to DNNs, such as randomized importance \cite{random}, rule lists \cite{interpretable}, partial dependence plots \cite{zhao2017causal, greedy, molnar_interpretable_nodate}, Shapely scores \cite{shapely, molnar_interpretable_nodate}, and Bayesian case based models \cite{been}. 

\section{Definitions}
We follow from previous work \cite{review} that a proper explanation should be both interpretable and complete.  By intepretable, we mean that the explanation should be understandable to humans.  That does not necessarily imply that the explanation must be in human-readable form, in fact, visual cues are well-understood by humans.  When we say that the methods must be complete, we mean the resulting explanation should be true to the model.  For example, while using a simplified model that is explainable (like a linear model) to fit the input to the output results in a nice explanation, it is not a true and complete representation of the internal concepts, representations, and decisions of the model.   

In this paper, we refer to \emph{inside} and \emph{outside} explanations for explaining DNNs.  When we refer to inside explanations, we are referring to the type of explanations that currently exist, that are catered towards AI developers and experts.  They are tailored to people \emph{inside} the field.  We encourage the development of \emph{outside} explanations that are interpretable, complete, and answer why questions.  They build trust not only to their technical developers, but also those \emph{outside} the technical scope that may use their technology without a technical background.  


\section{Current Limitations}
To show the strengths, benefits, and challenges of current explanatory approaches for opaque, DNN systems, we use a previously defined taxonomy  \cite{review} . The taxonomy consists of 3 classes.  The first class are systems that explain processing by looking at the relationships between the inputs and the outputs.  These include salience mapping \cite{cam, grad-cam}, decision trees \cite{deepred}, automatic rule-extraction \cite{rule-survey}, and influence functions \cite{influence}. 
The second class are systems that explain representation for DNNs either in terms of layers \cite{razavian, yosinski}, neurons \cite{netdissect} or vectors \cite{cavs}.  The final class is explanation-producing systems that look at attention-based visual question answers \cite{multimodal} or disentangled representations \cite{capsule} to create self-explaining systems.  

\begin{table}
  \caption{Strengths, Benefits, and Challenges of Current DNN XAI systems}
  \label{question-table}
  \centering
  \begin{tabular}{|c|p{4.2cm}|p{5cm}| }
    \hline 
    \bf{Method} & \bf{Questions it can answer} & \bf{Questions it cannot answer} \\
    \hline
    Processing & Why does this particular input lead to this particular output?  &  Why were these inputs most important to the output? How could the output be changed? \\
    \hline
    Representation & What information does the network contain?  & Why is a representation relevant for the outputs? How was this representation learned? \\
    \hline
    Explanation &  Given a particular output & What information contributed to this  \\
    producing &  or decision, how can the network explain its behavior?  & output/decision?  How can the network yield a different output/decision?  \\
    \hline
  \end{tabular}
\end{table}

When examining this taxonomy for policy purposes, the biggest shortcoming is that these systems cannot explain \emph{why}.  There are two types of questions that we should ask of a decision making algorithm:
\begin{enumerate}
    \item Why did this output happen?
    \item How could this output have changed?
\end{enumerate}

A summary of the types of questions that current DNN XAI systems can and cannot answer are in Table \ref{question-table}.  Explanation producing systems nearly answer the first question we would want to ask a decision making algorithm: why did this output happen?  But the problem is that their explanation may not be \emph{complete} and \emph{true} to the model's internal decisions and processes.  In order to illustrate the necessity of answering these questions, we proceed by  walking-through examples of an AI algorithm, a larger AI system, and illustrate problems with data to motivate why explanatory systems should strive to answer the preceding questions.  


\subsection{Societal needs for explanations}

Imagine you do not receive a loan, you would want to know what was the key attribute that limited the algorithm. You would want to know \emph{why} you were denied a loan. But further, you may also want a sensitivity analysis: what would you need to change to be able to get the loan.  There may be several possibilities.  For example, you may have received a loan if you made \$1,000 more per month; something you may be able to change in the future. However, other factors may be things you cannot control, such as the specific time you applied or your gender or ethnicity. So, in this case, we would like to have system that is able to explain why it decided to give or not a loan to each person. 

Moreover, consider again the AI system example mentioned earlier of a self-driving car involved in an accident. The first thing we would want to know is why the accident happened. In this case there are many algorithms interacting. Finding if there was a faulty component is extremely challenging, making it even more relevant for each part of the system to be able to explain its decisions. In the recent Uber accident where the vehicle struck and killed a pedestrian, detecting the root-cause of the accident took several weeks to uncover in the complex AI software system \cite{marshall_uber_2018}.

But the other, more challenging question we would want to ask is if the accident could have been avoided. This is a more difficult question than the previous, single algorithm question. In complex systems, an error could be local (caused by a single failure), or it could be caused by an inconsistency between parts working together. The latter is much more difficult to detect, diagnosis, and explain.  

In the Uber case, since the accident was deemed to be caused by a false positive\cite{lee_report:_2018} on the error detection monitoring the pedestrian, several explanations could provide evidence of how this could have been avoided.  Again, some inconsistencies are easier to fix than others (which may not be possible).  Perhaps the sensitivity on the error detection monitor should be decreased or increased.  Or perhaps the pedestrian would have been detected with higher certainty during the daytime, or if they were walking slower.  It is still left to question whether the training data was at fault, which introduces a new set of questions.  

\subsection{The risk of opaque models}
Generally explaining model behavior is not enough to build trust in these sorts of models. Another way these algorithms and systems can behave badly is due to a inconsistency in the training data and/or knowledge bases.  This does not necessarily mean that the training data is ``bad'' per say, but that there is a misalignment between the expected data and the actual training data used.  We have seen this recently with the Amazon recruiting algorithm \cite{dastin_amazon_2018}.  This algorithm was eventually disbanded because the results were extremely biased; since the algorithm had been trained on applicants data for the past 10 years (where males are dominant), it was teaching itself to choose male candidates.  Even if the algorithm was modified, there was no way to ensure it was unbiased.  Although this is an extremely compelling case for inquisitive explanatory systems, an even more persuasive case is for safety-critical tasks.  

Equally important, consider a machine learning classifier to diagnose breast cancer from an image, where the training set was carefully selected to be fairly close to a 50-50 split of breast cancer and non-breast cancer scans. Even if the classifier is very accurate, without having access to complete explanations to understand how decisions are made in the model, it is not certain that it is making decisions for the right reasons---the model may in fact, learn a feature it should not rely on despite predicting breast cancer very accurately. In \cite{leakage}, one classifier learned the resolution of the scanner camera, therefore predicting cancerous images from a high resolution very accurately. Figuring out this sort of data problem is extremely difficult. It requires either an attuned intuition of the model's inner workings or the model to be able to answer questions to do a fine-grain sensitivity analysis.



\section{Discussion and Conclusion}
As humans, we start to build trust by asking questions.  We should be able to judge the behavior of opaque DNN algorithms by asking similar questions as we would ask of a person’s behavior in similar circumstances. The key idea here is that explainability exceeds transparency and interpretability, to empower the public to understand the decisions and underlying mechanisms. 

We have focused mainly on explainable DNN algorithms, but when a DNN algorithm is a part of a larger system,  explainability is not enough.  Explainability does not necessary imply that complex systems are accountable and responsible; this may have to be tackled with other requirements.  For example, a system can provide an ``outside'' explanation without addressing who or what is responsible and why. At the same time, an explainable system may be transparent without being receptive to human feedback. In future work, we will examine how explainability may interact with other parts of  systems (including the human operator or user) to produce systems that can be augmented and learn from feedback.



But in order to truly trust AI systems, people will not only need to feel confident that they understand certain how decisions are made, but also that they have recourse.  If a person disagrees with a system's output, they should be empowered and able to change the system. Designing explainable AI is important, but only when opaque systems are auditable, explainable, answer questions, and interpret feedback will we be confident enough to trust their decision-making.  




\bibliographystyle{plain}
\bibliography{main}






\end{document}